\documentclass[runningheads]{llncs}
\usepackage{graphicx}
\usepackage{booktabs}
\usepackage{color}
\usepackage{multirow}
\usepackage{amssymb}
\usepackage[colorlinks,linkcolor=blue,anchorcolor=blue,citecolor=blue,urlcolor=blue]{hyperref}
\usepackage{array}
\newcolumntype{L}[1]{>{\raggedright\let\newline\\\arraybackslash\hspace{0pt}}m{#1}}
\newcolumntype{C}[1]{>{\centering\let\newline\\\arraybackslash\hspace{0pt}}m{#1}}
\newcolumntype{R}[1]{>{\raggedleft\let\newline\\\arraybackslash\hspace{0pt}}m{#1}}
\begin{document}
\title{MVB: A Large-Scale Dataset for Baggage Re-Identification and 
Merged Siamese Networks}
\titlerunning{MVB: A Large-Scale Dataset for Baggage ReID.}
%
\author{Zhulin Zhang\inst{1} \and
Dong Li\inst{1} \and
Jinhua Wu\inst{1} \and
Yunda Sun\inst{1} \and
Li Zhang\inst{1}}
\authorrunning{Z. Zhang et al.}
%
\institute{Nuctech AI R\&D Center, 100084, Beijing, China\\
\email{\{zhangzhulin, li.dong, wujinhua, sunyunda, zhangli\}@nuctech.com}}
\maketitle              
\begin{abstract}
In this paper, we present a novel dataset named MVB (Multi View Baggage) for baggage ReID task which has some essential differences from person ReID. The features of MVB are three-fold. First, MVB is the first publicly released large-scale dataset that contains 4519 baggage identities and 22660 annotated baggage images as well as its surface material labels. Second, all baggage images are captured by specially-designed multi-view camera system to handle pose variation and occlusion, in order to obtain the 3D information of baggage surface as complete as possible. Third, MVB has remarkable inter-class similarity and intra-class
 dissimilarity, considering the fact that baggage might have very similar appearance while the data is collected in two real airport environments, where imaging factors varies significantly from each other. Moreover, we proposed a merged Siamese network as baseline model and evaluated its performance. Experiments and case study are conducted on MVB.

\keywords{Dataset  \and Re-Identification \and Siamese Networks.}
\end{abstract}
\section{Introduction}
At international airports, baggage from flights normally need to be scanned by security check devices based on X-ray imaging due to safety issues and customs declaration. To increase the customs clearance efficiency, X-ray security check devices have been deployed in BHS (Baggage Handling System) at many newly constructed airports. After flight arrivals, all check-in baggage will go through security check devices, which are connected with conveyor of BHS. Therefore, the X-ray image of each baggage is generated and inspected before baggage claim. Currently, the common practice is attaching RFID (Radio Frequency Identification) tags onto interested baggage right after security check devices, in order to indicate the baggage to be further manually unpacked and inspected. As passengers claim interested baggage with RFID tags and carry it to RFID detection zone, alarms will be triggered. 

Nevertheless, RFID tag detection has certain drawbacks. First, tags might fall off in the process of transfer. Certain passengers might also deliberately tear off tags in order to avoid inspections. The loss of tags will directly result in detection failures of interested baggage. Second, tagging need to be conducted by manpower or certain equipment, which causes additional cost together with the tag itself and might affect customs clearance rate. Moreover, baggage of metal material surface will interfere with detection signal of RFID tags, thus it also leads to false negative cases. 

Considering these defects, a security inspection approach that requires no physical tags will show great advantages in avoiding detection miss and metal interference, reducing costs, and increasing efficiency. An approach based on baggage appearance images is thus employed. Concretely, images of baggage appearance will be captured at BHS and bundled with inspection information before baggage claim. While passengers carrying the baggage and entering the customs checkpoint, i.e. the area for customs declaration and security check before leaving the airport, the appearance image will be taken again. These checkpoint images will be analyzed by comparing with those taken at BHS to identify whether certain baggage is of interest. Practically, passengers often place feature items such as stickers or ropes on baggage, which can serve as cues in distinguishing baggage, thus each baggage could be to be unique within certain time interval. Since the baggage is re-identified cross cameras, the process is referred as baggage ReID later in this paper.

Similar to the person ReID~\cite{ref_proc1}, the baggage ReID task also faces challenges such as object occlusion, background clutter, motion blurring and variations of lighting, pose, viewpoint, etc. Particularly, some of these aspects are even more challenging for baggage ReID. For instance, the baggage pose often differs between images captured at BHS and checkpoint, as well as per each baggage. It brings extra difficulties for applying part-based person image retrieval approaches~\cite{ref_proc2,ref_proc16} to baggage ReID, since pedestrian in video surveillance mostly remains canonical standing/walking pose. Meanwhile, similar to vehicle ReID~\cite{ref_proc3,ref_proc4}, baggage images from different view-points vary much more than the case of person ReID. Furthermore, it is not uncommon that many baggage has very similar appearance thus are less distinctive compared with person. All these characteristics make baggage ReID a uniquely challenging task.

Recent years, research and application in computer vision have seen great development, especially with the help of deep learning. An important enabling factor of the rapid development of deep learning is the availability of large scale datasets~\cite{ref_proc5,ref_proc6,ref_proc10}. Taking person ReID as example, datasets such as Market-1501~\cite{ref_proc7}, MARS~\cite{ref_proc8}, CUHK03~\cite{ref_proc9}, etc., have contributed to improving the state-of-the-art performance continuously~\cite{ref_proc16,ref_proc19}. These large-scale datasets played a key role to evolve the person ReID task from lab problem to real-world industrial application.

In this paper, a large-scale baggage ReID dataset called MVB (Multi View Baggage) is proposed. First, as a large-scale image dataset, MVB consists of 4519 baggage identities and 22660 annotated hand-drawn masks and bounding boxes, as well as surface material labels. Second, all baggage images are captured by specially-designed multi-view camera system to handle pose variation and occlusion. The multi-view images contribute to obtaining 3D information of baggage surface as complete as possible, which is crucial to the ReID problem, since there could be notably different textures on specific area of baggage. Third, in real scenario at airports, the imaging factors like lighting, background, viewpoint, motion, etc., are quite different between BHS and checkpoint, making the baggage ReID task of our dataset tend to be a cross domain problem, which is more challenging and inspiring. Moreover, baggage might have very similar appearance thus are hardly distinctive. These aspects make our dataset have remarkable inter-class similarity and intra-class dissimilarity which domain adaptation approach~\cite{ref_proc17,ref_proc18} in person ReID could be applied. To the best of our 
\begin{figure}
\centering
\includegraphics[width=0.9\textwidth]{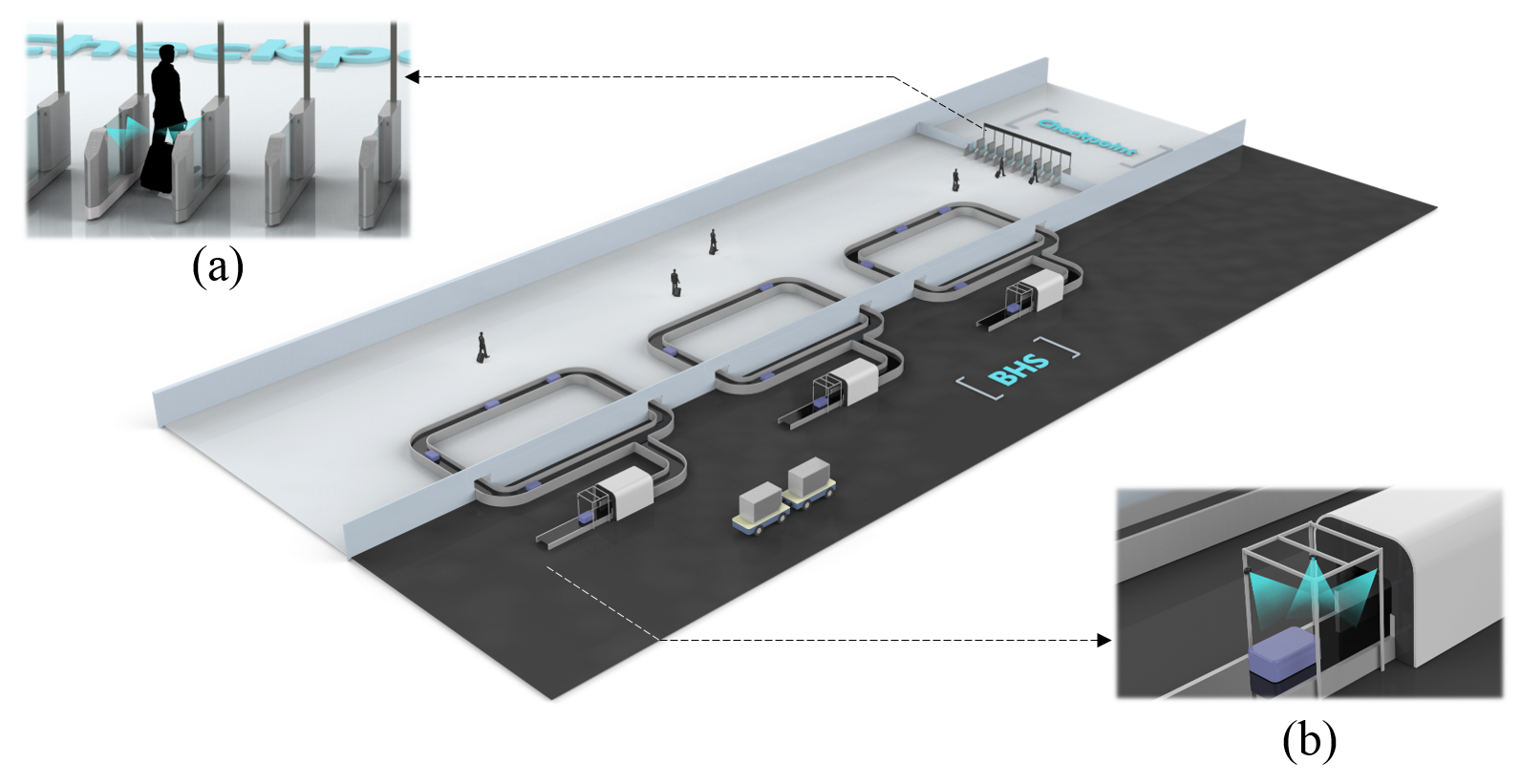}
\caption{Baggage ReID application and multi-view camera system at: (a) checkpoint (b) BHS.} \label{fig1}
\end{figure}
knowledge, MVB is the first publicly available baggage ReID dataset, which will enable utilizing deep learning methods on baggage ReID and benefit research and application on general object ReID tasks. Additionally, we also propose baseline models using merged Siamese network with ablation study to understand how baggage ReID performance benefit from features like self-attention, hard example mining, foreground mask, etc.

This paper is organized as follows. In Section 2, MVB dataset will be introduced in detail. Task and evaluation method on MVB will be given in Section 3. Baseline models and corresponding experiment results will be shown in Section 4 and 5. In Section 6, a short conclusion will be summarized.

\section{Dataset}
\subsection{Raw Data Collection System}
As raw data, images containing baggage are all captured at an international airport. The baggage ReID application is illustrated in Fig.~\ref{fig1}. The data collection process can be divided to two stages, i.e. BHS and checkpoint, both have multi-view image capture system deployed. 

In BHS stage, after unloaded from landed airplanes, baggage is put on BHS conveyor and transferred to a security check device for X-ray scan in sequence. At the entrance of the device, a portal frame is set up over the conveyor. In order to get 3D information of baggage appearance as complete as possible, three cameras were placed on different position of the frame to capture multi-view images: right-front, top, and left-back respectively. These cameras receive the trigger signal as baggage passes by and take three images simultaneously. As the baggage being scanned by the device next to the frame, the generated X-ray image can be inspected by staff or algorithm in real-time, then the information of whether certain baggage is of interest is bundled with the multi-view images taken by the cameras.

The second stage for capturing multi-view images is at the checkpoint for customs clearance. According to procedure of customs clearance, passengers along with baggage are required to pass through gate at checkpoint after baggage claim. The checkpoint usually contains several gates. At each gate, four cameras are embedded for multi-view image capturing. Two pairs of cameras are located near the exit and entrance of the gate at each side, taking images against and along the passenger moving direction respectively. The two pairs of cameras are triggered in proper order to adapt many passenger actions such as pushing a baggage cart, dragging/pushing a mobile suitcase, etc. The intention of embedding four cameras is trying to capture baggage with different possible poses, such as lying on baggage cart and standing on ground, considering the fact that in some view the particular baggage might be heavily occluded by person or other baggage.

\subsection{Data Annotation}
Based on the multi-view image capturing system, raw image data were collected at an airport from actual flight during several days. In real case, a baggage ReID pipeline consists of two sequential steps, baggage detection and baggage retrieval. In this paper, the detection step is not considered in the pipeline of baggage ReID for mainly two reasons. First, we have trained Faster-R-CNN~\cite{ref_proc11} based object detection models using annotated bounding boxes on full-sized images, it showed that using detection result for retrieval task has almost the same performance compared to using ground truth. Second, a baggage can be identified means it has at least one valid baggage image taken at BHS and checkpoint respectively. Since there could be many hold-on baggage also appeared in checkpoint image besides check-in baggage, the annotation for detection might bring the dataset many irrelevant baggage which are unable to identify. Therefore, we refer baggage retrieval as baggage ReID in our paper.
\begin{table}
\centering
\caption{Annotation Statistics.}\label{tab1}
\begin{tabular}{ C{2cm} C{3cm} C{3cm} C{4cm} }
\toprule
 & \#Baggage Images & \#Full-sized Images & Average Views per Identity\\
\midrule
BHS & 13028 & 13028 & 2.88\\
Checkpoint & 9632 & 9237 & 2.13\\
Overall & 22660 & 22265 & 5.01\\
\bottomrule
\end{tabular}
\end{table}

The annotation process can be described as follows. Images taken at BHS and checkpoint would be annotated if there is a valid baggage. Valid baggage denotes that one integrated surface of baggage is exposed at checkpoint or more than 50\% of baggage surface is exposed at BHS. Each mask is a hand-drawn polygon and each corresponding bounding box is then cropped as minimum enclosing rectangle of annotated mask. Because there are four camera views at checkpoint and three camera views at BHS, the first annotation for ReID is to couple the same baggage separately based on time. The second step is finding the same identity between checkpoint and BHS, which is quite a time-consuming work. Therefore, a ReID model is trained based on a few identities and computed the scores of similarity between baggage at BHS and checkpoint, the ground-truth identity would be much easier to locate based on ranking. At last, the annotator confirms that each identity consists of images from BHS and checkpoint.

MVB consists of 4519 baggage identities and 22660 bounding boxes. Each identity is examined to be unique. For each bounding box, mask of baggage is also given as annotation information. 22660 baggage images (13028 at BHS, 9632 at checkpoint) are cropped from 22265 full-sized images (13028 at BHS, 9237 at checkpoint). Most identities have three baggage images taken at BHS. The number of baggage images at checkpoint gate for each identity fluctuates more. Most frequent occurrence of missing baggage image from certain view at BHS is due to missed camera capture, while at checkpoint is more often due to serious occlusion caused by passenger body parts or cloth, baggage cart, other baggage on cart or on ground. On average, each baggage identity has respectively 2.88 and 2.13 baggage images at BHS and checkpoint. The statistics of annotation is listed in Table~\ref{tab1}.

For better baggage ReID evaluation, the dataset has also provided the attribute annotation of baggage surface material. The attribute labels of four categories are: hard (metal, plastic, etc.), soft (fabric, leather, etc.), paperboard and others (protective cover, etc.). Table~\ref{tab2} showed the sample baggage images and label distributions.

\begin{table}
\centering
\caption{Surface Material Annotation.}\label{tab2}
\begin{tabular}{ C{2.5cm} C{2cm} C{2cm} C{2cm} C{2cm} }
\toprule
Categories & Hard & Soft & Paperboard & Others\\
\midrule
\#Identities & 2767 & 1120 & 198 & 434\\
Sample Image & 
\includegraphics[width=1cm]{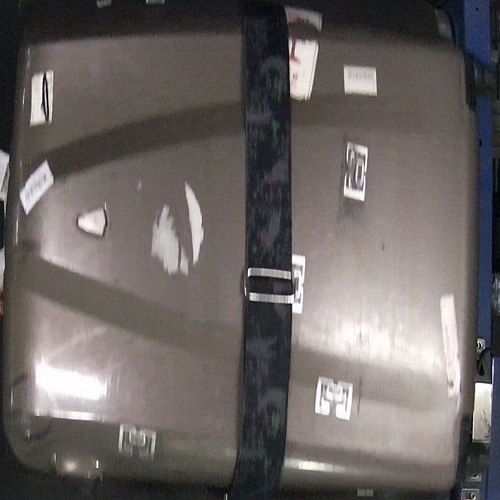} &
\includegraphics[width=1cm]{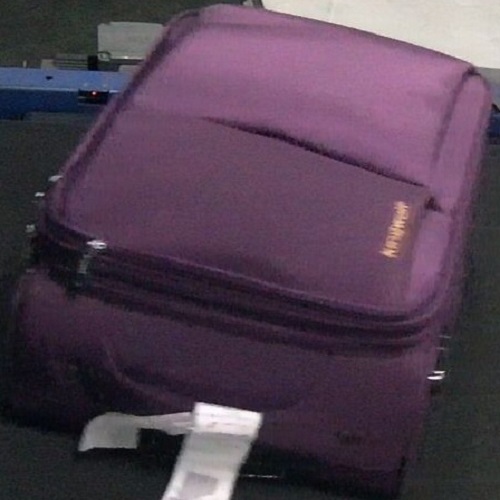} &
\includegraphics[width=1cm]{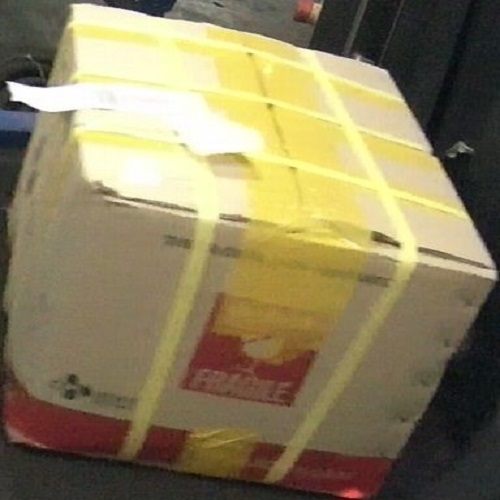} &
\includegraphics[width=1cm]{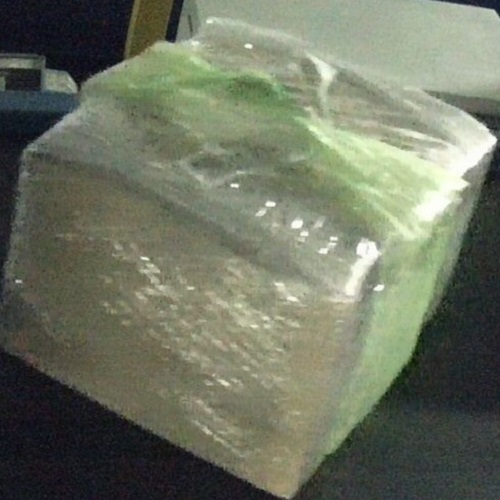}\\
\bottomrule
\end{tabular}
\end{table}

\subsection{Dataset Characteristics}
In MVB dataset, each identity of baggage can be regarded as an individual class containing several images taken at BHS and checkpoint together. It is necessary to point out the characteristics of inter-class similarity and intra-class dissimilarity. For inter-class similarity, we have to admit that some baggage is naturally very hard to distinguish from each other according to their appearance, even more difficult than the case in person ReID. For instance, Table~\ref{tab3} gives two baggage that looks very similar but actually has different identities. The cues to distinguish them are hiding in detail of images. Meanwhile, the images of BHS and checkpoint are substantially different. The intra-class dissimilarity aspects are listed in Table~\ref{tab4}.

\noindent
\textbf{Background:} as most of images in Table~\ref{tab4} indicate, baggage images have quite different backgrounds between BHS and checkpoint. In BHS images, background mainly consists of black conveyor belt and security check device entrance. Meanwhile in checkpoint baggage images background varies from passenger body parts, clothes, baggage cart, floor, etc.

\noindent
\textbf{Occlusion:}other baggage on cart can easily lead to heavy occlusion in checkpoint image as shown in Table~\ref{tab4}d, while checkpoint image might be also partially invisible in BHS image because surface is at bottom, which corresponds to the case in Table~\ref{tab4}c.

\noindent
\textbf{Viewpoint and pose:}they are essentially unlike due to different locations of cameras, and baggage can be in various poses such as Table~\ref{tab4}a showed.

\noindent
\textbf{Lighting:}lighting conditions at BHS and checkpoint are not the same which often leads to color and reflection differences. For instance, Table~\ref{tab4}b displays obviously different color characteristic at BHS and checkpoint.

\noindent
\textbf{Motion blur:}as passengers walking through checkpoint gate at different speed, motion blur makes baggage image to be less distinctive, as shown in Table~\ref{tab4}e.

All these above factors make baggage ReID on MVB a challenging and inspiring task between different domains. 
\begin{table}
\centering
\caption{Samples of inter-class similarity on MVB. Images in each row are from one identity.}\label{tab3}
\begin{tabular}{ C{0.5cm} C{1.8cm} C{1.8cm} C{1.8cm} C{1.8cm} C{1.8cm} }
\toprule
 & \multicolumn{2}{c}{Checkpoint} & \multicolumn{3}{c}{BHS}\\
 & View1 & View2 & View1 & View2 & View3\\
\midrule
a &
\includegraphics[width=1cm]{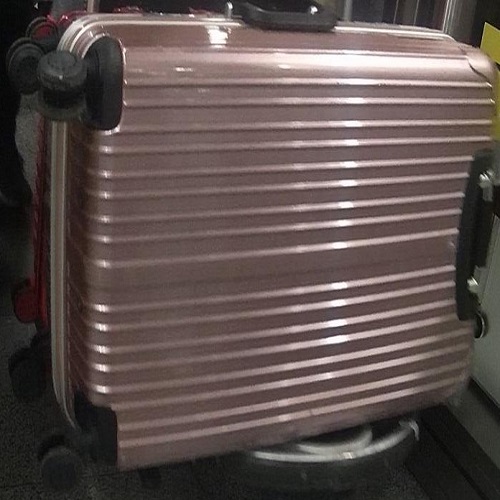} &
\includegraphics[width=1cm]{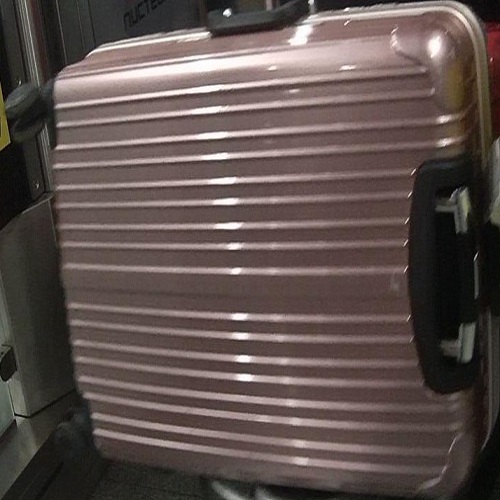} &
\includegraphics[width=1cm]{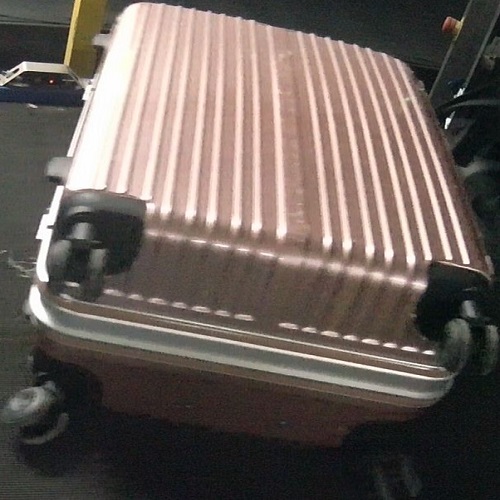} &
\includegraphics[width=1cm]{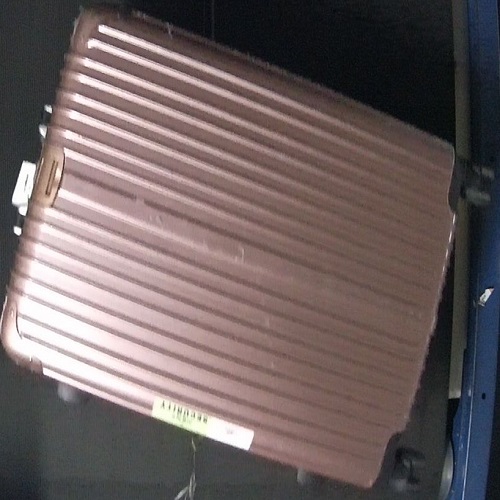} &
\includegraphics[width=1cm]{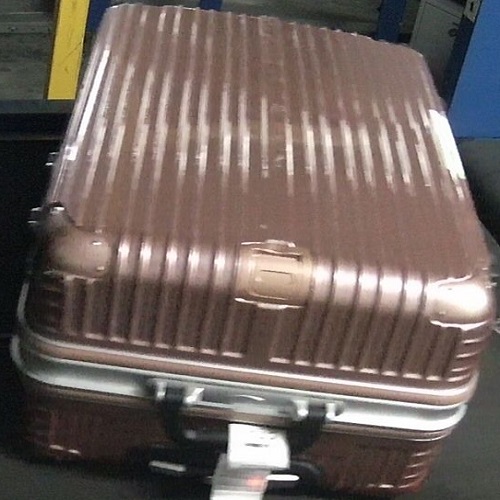}\\
b &
\includegraphics[width=1cm]{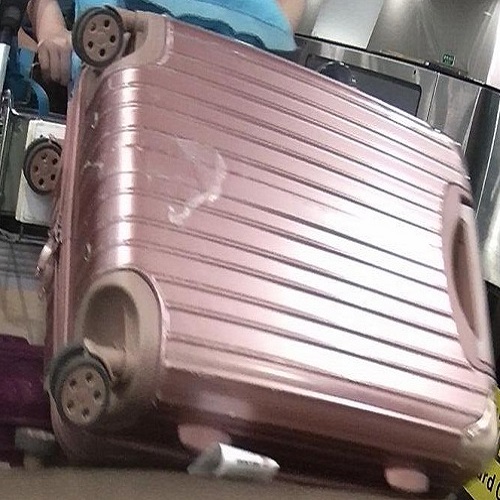} &
\includegraphics[width=1cm]{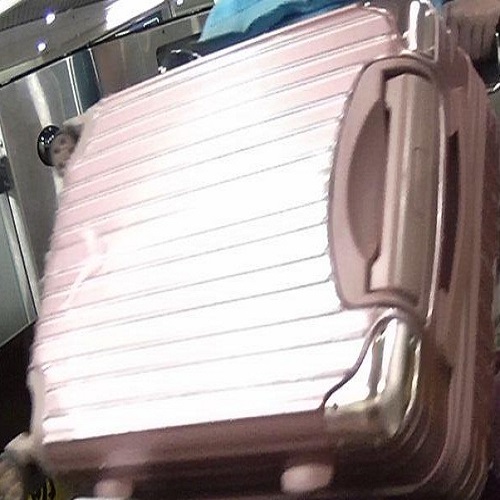} &
\includegraphics[width=1cm]{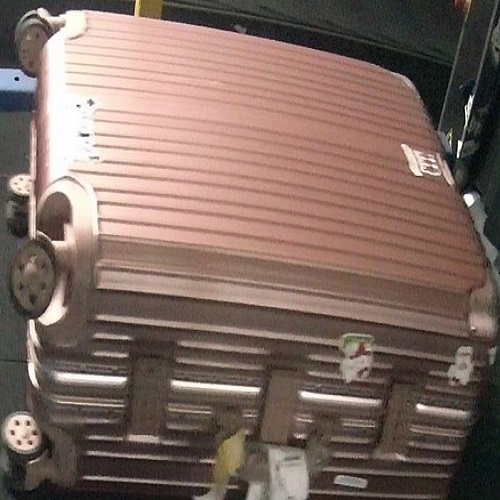} &
\includegraphics[width=1cm]{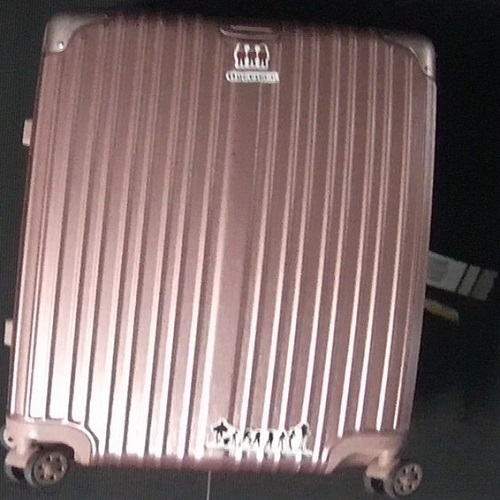} &
\includegraphics[width=1cm]{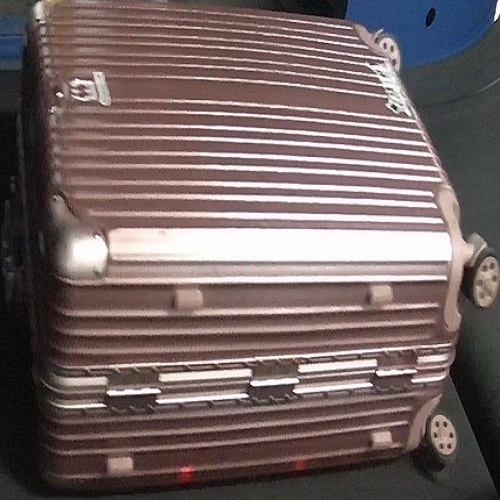}\\
\bottomrule
\end{tabular}
\end{table}

\begin{table}
\centering
\caption{Samples of intra-class dissimilarity on MVB. Blank cell indicates corresponding view image is not valid. Images in each row represent the same identity.}\label{tab4}
\begin{tabular}{ C{0.5cm} C{1.8cm} C{1.8cm} C{1.8cm} C{1.8cm} C{1.8cm} C{1.8cm} C{1.8cm} }
\toprule
 & \multicolumn{4}{c}{Checkpoint} & \multicolumn{3}{c}{BHS}\\
 & View1 & View2 & View3 & View4 & View1 & View2 & View3\\
\midrule
a &
\includegraphics[width=1cm]{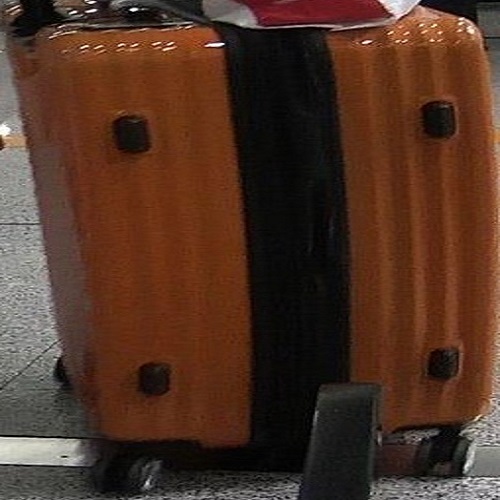} &
\includegraphics[width=1cm]{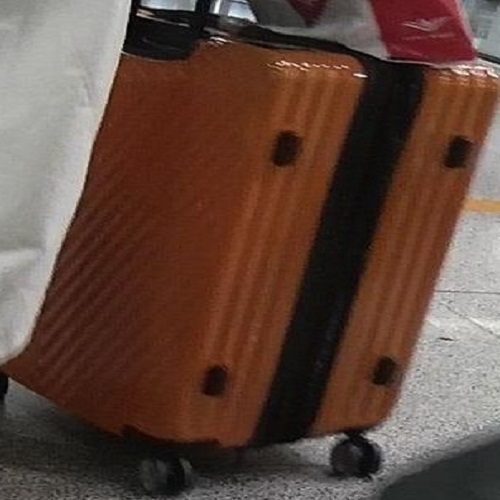} &
\includegraphics[width=1cm]{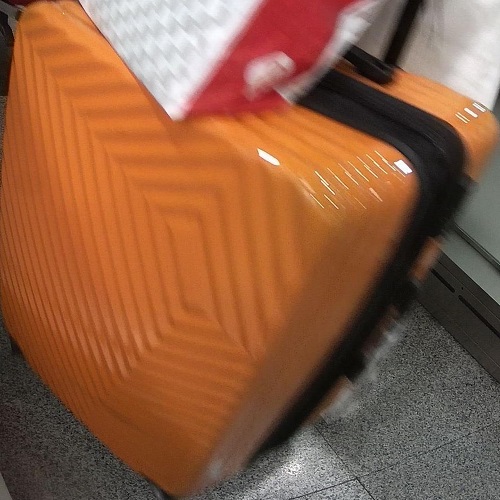} &
\includegraphics[width=1cm]{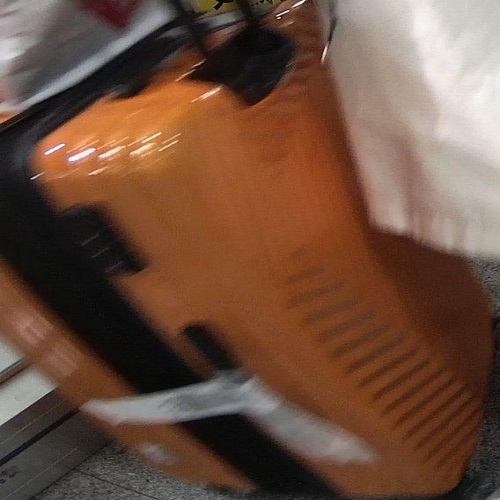} &
\includegraphics[width=1cm]{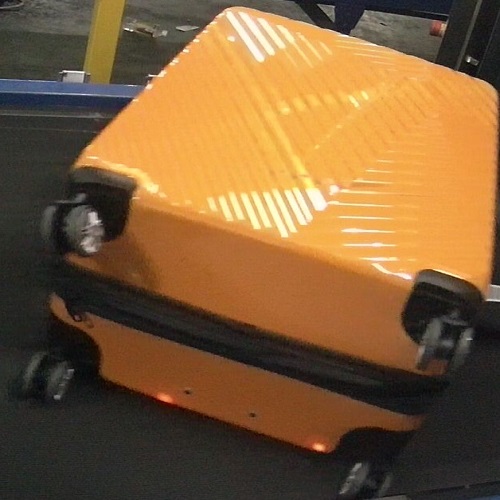} &
\includegraphics[width=1cm]{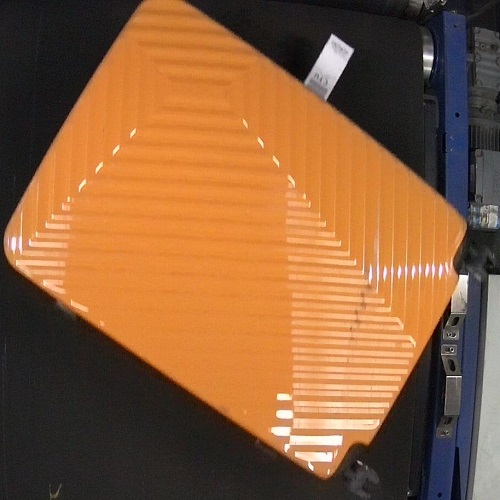} &
\includegraphics[width=1cm]{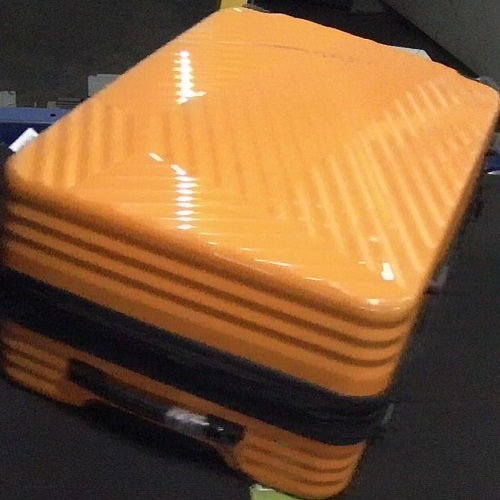} \\
b &
\includegraphics[width=1cm]{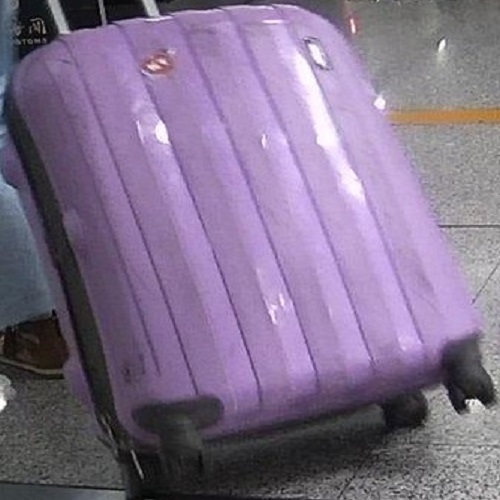} &
\includegraphics[width=1cm]{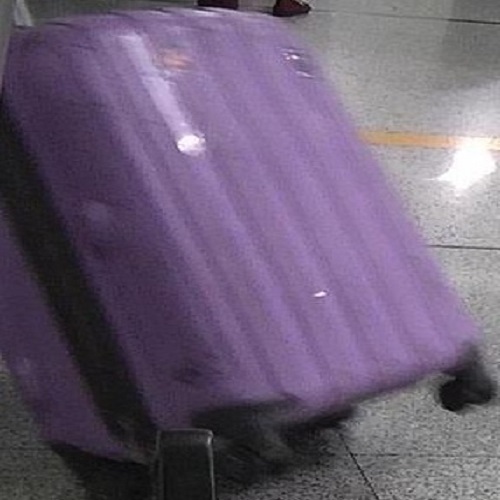} &
\includegraphics[width=1cm]{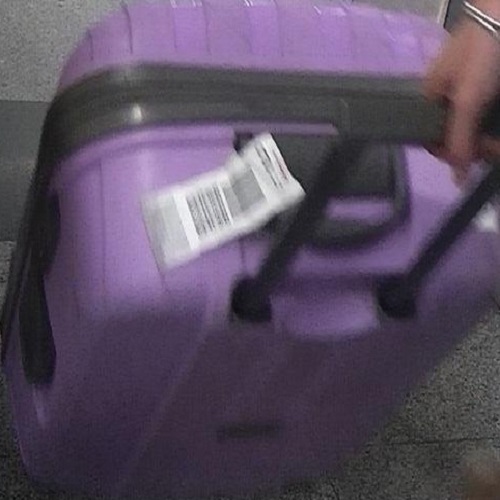} &
\ &
\includegraphics[width=1cm]{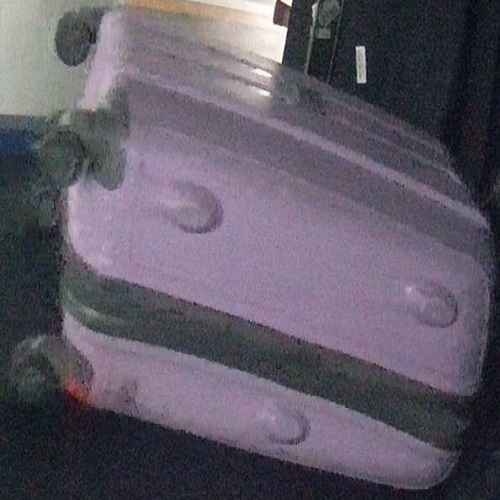} &
\includegraphics[width=1cm]{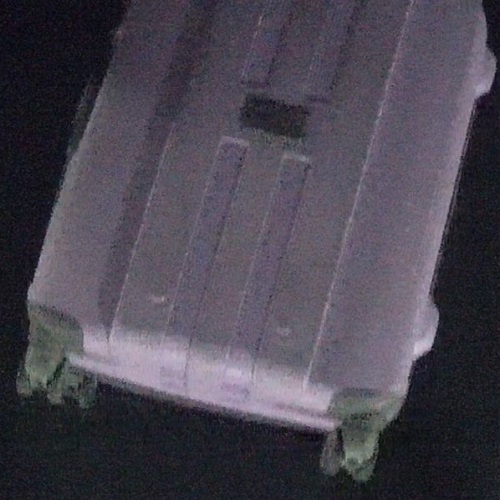} &
\includegraphics[width=1cm]{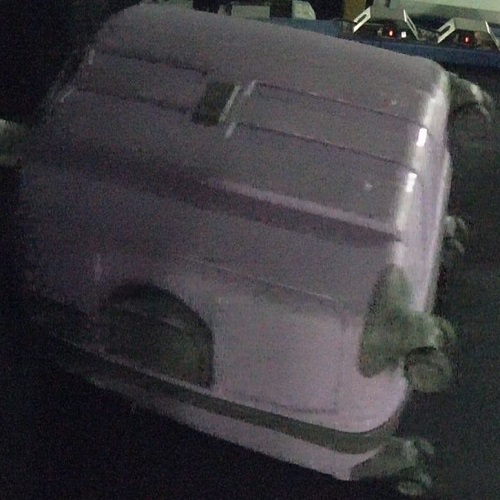} \\
c &
\ &
\ &
\includegraphics[width=1cm]{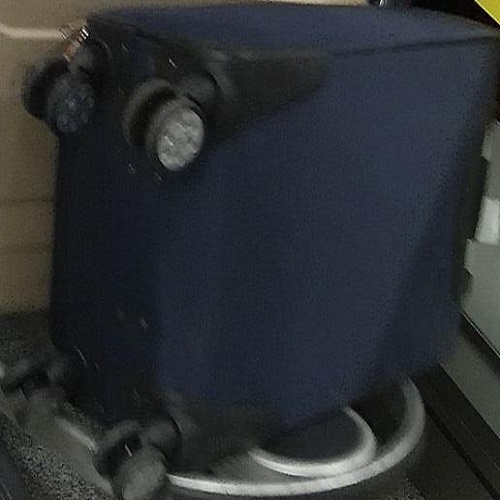} &
\includegraphics[width=1cm]{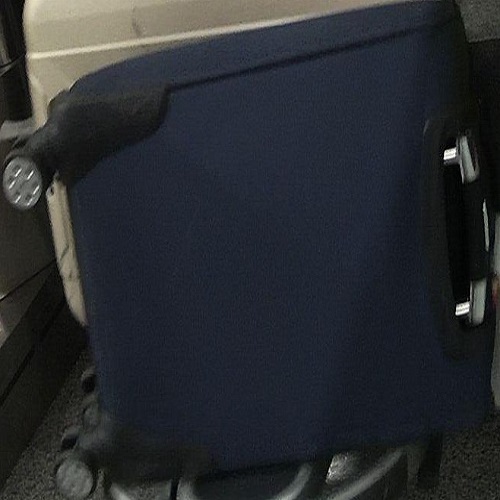} &
\includegraphics[width=1cm]{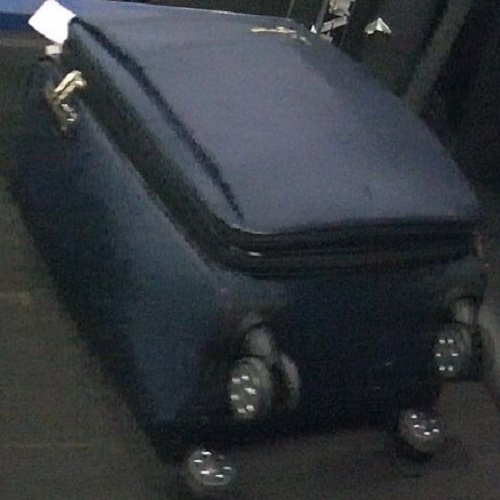} &
\includegraphics[width=1cm]{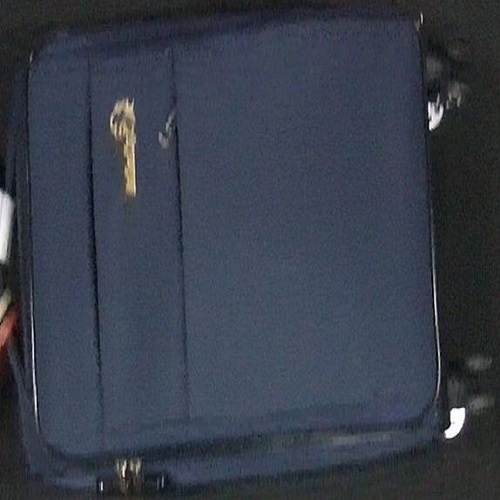} &
\includegraphics[width=1cm]{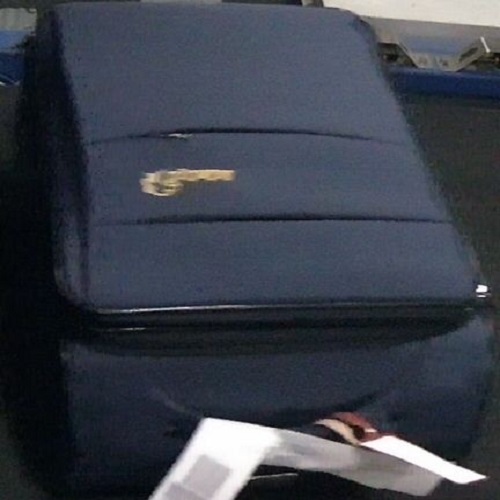} \\
d &
\ &
\ &
\includegraphics[width=1cm]{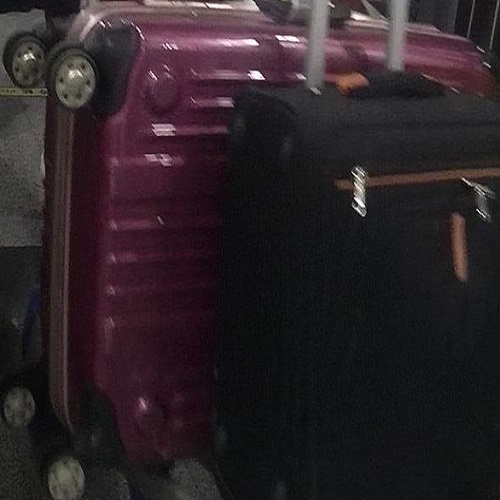} &
\includegraphics[width=1cm]{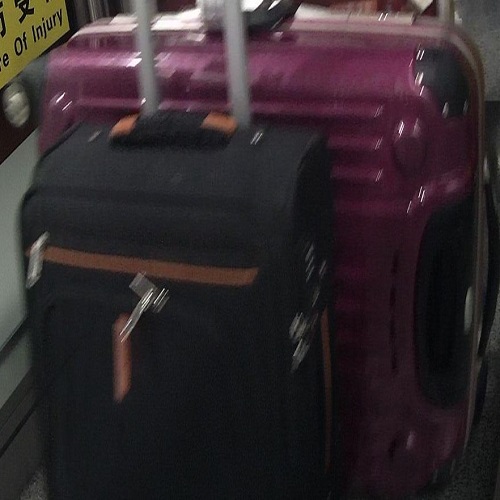} &
\includegraphics[width=1cm]{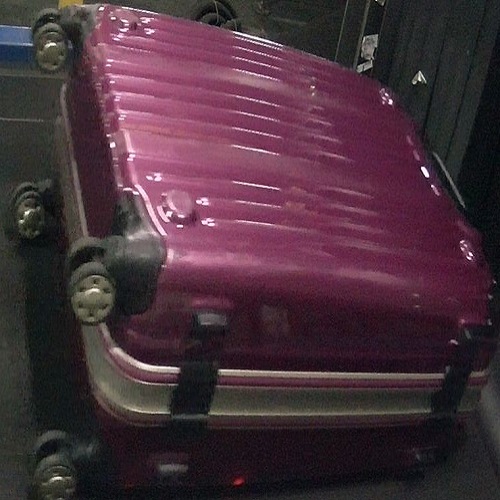} &
\includegraphics[width=1cm]{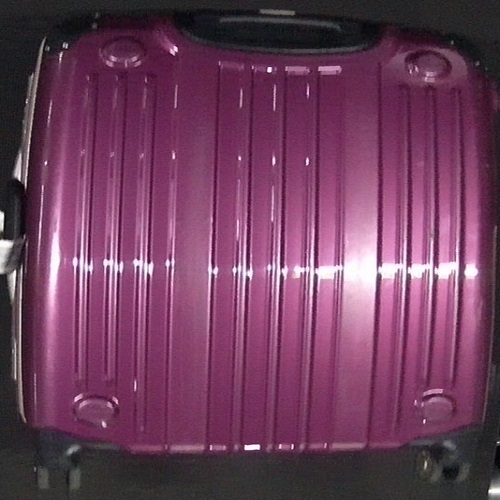} &
\includegraphics[width=1cm]{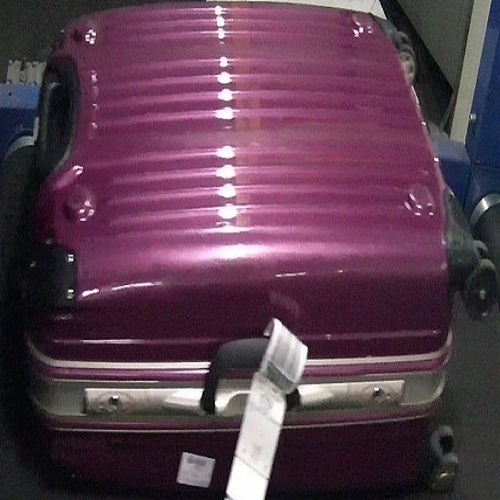} \\
e &
\ &
\ &
\includegraphics[width=1cm]{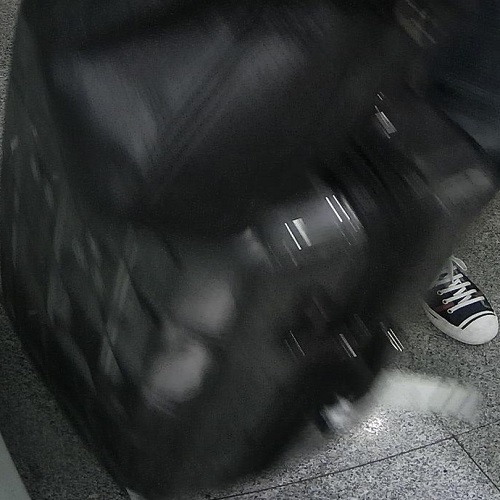} &
\includegraphics[width=1cm]{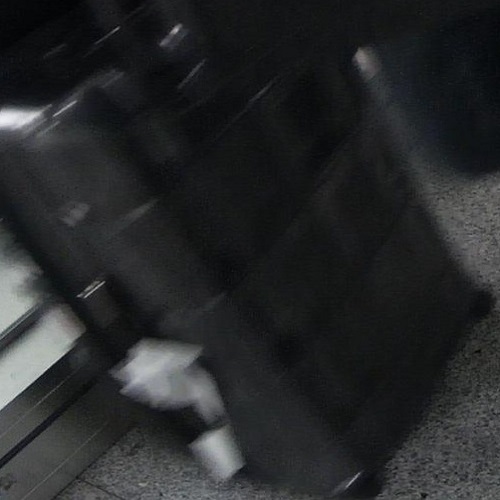} &
\includegraphics[width=1cm]{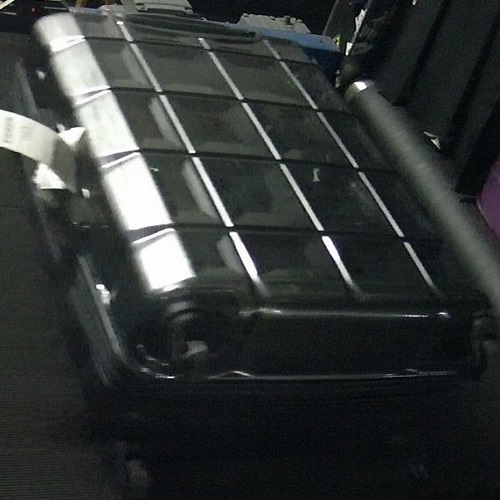} &
\includegraphics[width=1cm]{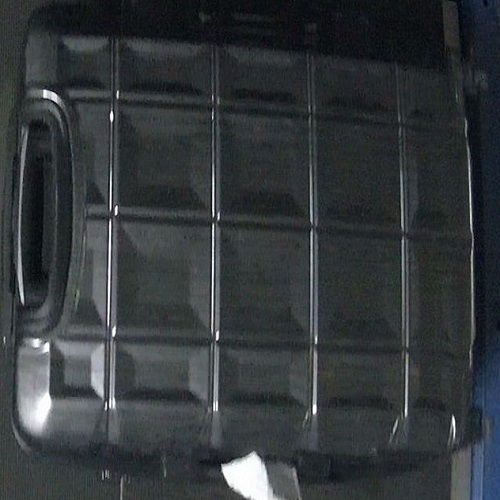} &
\includegraphics[width=1cm]{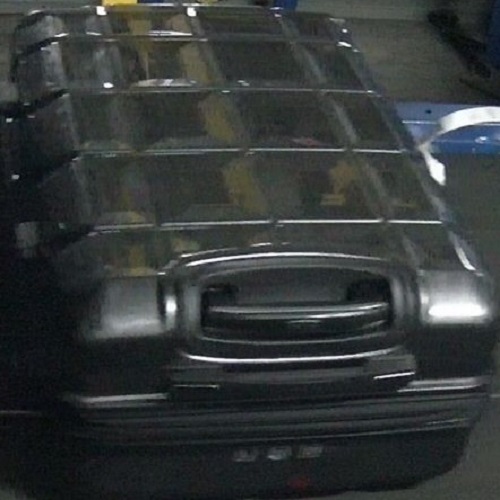} \\
\bottomrule
\end{tabular}
\end{table}

\section{Task and Evaluation Metric}
The task of baggage ReID on MVB is to assign a baggage identity to a given probe by searching among gallery. In baggage ReID task on MVB, definition of probe and gallery are not exactly the same as person ReID based on application scenario. Due to the cross domain characteristic, probe and gallery are naturally separated. Specifically, baggage will be taken appearance images at BHS before the domain is transferred from BHS to checkpoint. Baggage will be detected in checkpoint domain and then searched in BHS domain. Therefore, baggage images captured at checkpoint and BHS are defined as probe and gallery respectively. During test, gallery images from different views of the same identity are supposed to be treated as a whole in identifying whether a probe corresponds to a certain identity. Specifically, for each probe, inference result is supposed to be a possibility rank of all identities rather than all gallery images. Information of which gallery images belong to the same identity is given in test set, which can be easily obtained due to the same trigger signal introduced in Section 2.1.

Among 4519 identities in MVB, 500 identities randomly selected from all identities are reserved for test, while all the rest 4019 identities can be used for training. For the 500 identities test set, there are 1052 probe images and 1432 gallery images. Each probe image will be matched with the 1432 gallery images and a 500 id-length result vector will be output, indicating the sorted baggage under certain similarity metric. How to incorporate matching results of probe with multiple gallery images within an identity to single similarity value is left to be determined by dataset user. CMC (Cumulated Matching Characteristics) is adopted as evaluation metric to measure the performance of baggage ReID on MVB since there is only one ground-truth identity among gallery of 500 identities. In this paper, CMC at rank1 till rank3 will be evaluated.

\section{Baseline Method}
One nature of dataset MVB lies in large number of identities yet limited number of images within each identity, which might make classification scheme less feasible. In this paper, verification scheme using deep neural network is adopted for baggage ReID task. A basic Siamese network and a merged Siamese network are introduced in Section 4.1 and 4.2 respectively.

\subsection{Basic Siamese Network}
Siamese network is originally put forward for verification of signatures~\cite{ref_proc12}. Our basic Siamese network takes in two input images, processes these inputs using the same network architecture sharing parameters and subsequently produces two feature vectors. Ideally the distance under certain metric between the two output vectors indicates whether the two input vectors are from the same identity or not.

In the basic Siamese network adopted in our baggage ReID task, VGG16~\cite{ref_proc13} is used as backbone model to extract output feature vectors for input probe and gallery image. Euclidean distance between these two feature vectors is further calculated as similarity metric. In training phase, contrastive loss is adopted as loss function, with the intention of pushing Euclidean distance of same identity feature vectors near while pulling different identity feature vectors apart.

\subsection{Merged Siamese Network}
\begin{figure}
\centering
\includegraphics[width=0.9\textwidth]{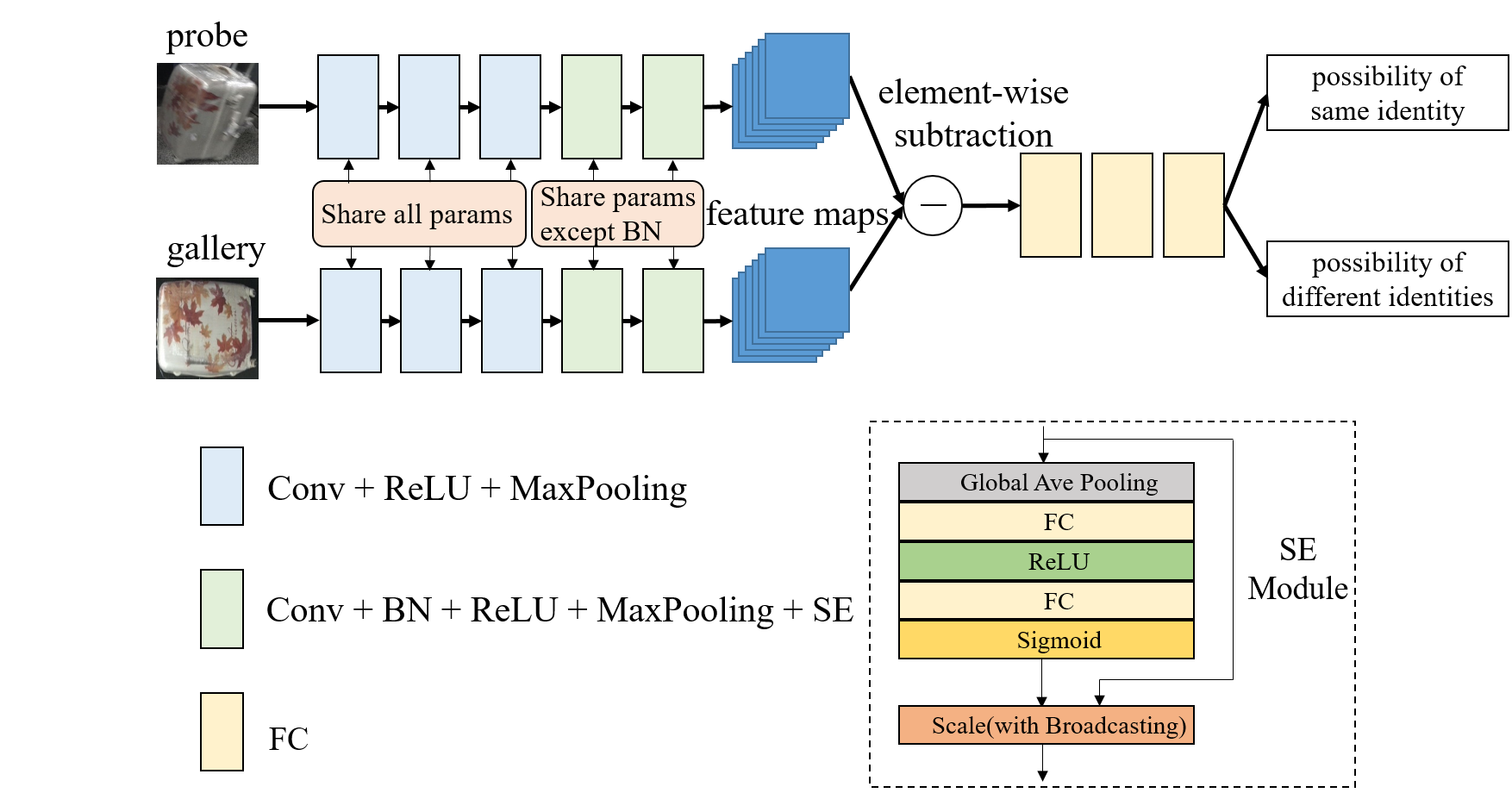}
\caption{Architecture of merged Siamese network.}\label{fig2}
\end{figure}

Our proposed merged Siamese network treats the verification problem as binary classification, as shown in Fig.~\ref{fig2}. Concretely, feature maps for probe and gallery image are extracted after the last convolution layer of VGG16. Then an element-wise subtraction layer is conducted on the feature maps of two paths and the output is fed into the fully connected layers for binary classification. The classification part of network generates possibility of whether probe and gallery images are from the same baggage identity, cross-entropy loss is adopted as loss function in training.

Compared with the basic Siamese network, feature maps extracted after the last convolutional layer contain more spatial information for further merging. The motivation behind element-wise subtraction lies in that by such operation co-located similar features at feature maps are suppressed while prominent dissimilar features are emphasized, meanwhile the spatial information is reserved. The subtraction output is further fed into binary classification network with fully connected layers to learn a similarity metric, which has more nonlinearity compared with Euclidean distance metric. Given the remarkable difference between probe domain and gallery domain, batch normalization~\cite{ref_proc15} is added in Conv4 and Conv5, and it should be noted that all parameters except batch normalization are shared for feature extraction of probe and gallery. 

Considering that channels of feature map might have different representation power, a channel-wise module based on Squeeze-and-Excitation (SE)~\cite{ref_proc14} is inserted after pooling layer in Conv4 and Conv5, aiming at learning weighted inter-channel relationship explicitly. The motivation behind Squeeze-and-Excitation module is to assign higher weight for more informative feature channels meanwhile lower weight for less informative ones.  In baggage ReID problem specifically, feature channels can be reasonably assumed to be informative to different extent. For instance, channels in which more activated features are from background rather than baggage should be suppressed. Since no external information other than feature itself is needed, channel-wise attention in form of SE can be viewed as a self-attention mechanism. The parameters for SE module are shared between probe and gallery.

\section{Experiments}
The basic and merged Siamese networks that introduced in Section 4 are evaluated on MVB dataset. 4019 identities and 500 identities are employed for training and test respectively. Both Siamese networks are finetuned from a pretrained VGG16 model, setting parameters in Conv1 and Conv2 to be frozen. Training is performed on 4$\times$NVIDIA Tesla P100 GPUs for 50k iterations with a minibatch of 128 image pairs. All probe and gallery images are resized to 256$\times$256 and then randomly cropped to 224$\times$224 in training phase.

For generating the pair data for Siamese network training, all positive pairs, i.e. pair of baggage images with the same identity, among 4019 identities are used as training data, meanwhile negative training pairs are randomly sampled among different identities, forming a training set balanced in positive and negative labels. The merged Siamese network is firstly trained on this balanced training set with a few epochs. Then the output model is utilized to inference each probe among 300 identities randomly sampled from 4019 identities for hard example mining. False positive pairs with high probability are filtered as supplement negative pairs then added to training set. The amount ratio of positive and negative pairs in the augmented training set is roughly 1:2, and total number of pairs is around 75k.

Training and evaluation are conducted on original baggage images and masked baggage images respectively. The masked baggage image is generated in a simple manner by keeping the pixel value inside the annotated polygon area and setting pixel value outside polygon area as zero. 

At test time, distance and possibility are inferenced between probe and each image in gallery. For each identity, mean of nearest two distances is regarded as the distance between probe and corresponding identity. Similarly, in classification scheme, mean of highest two possibilities within each identity is regarded as the possibility of same identity. For the minority identities with only one gallery image, computing mean value is replaced with the only distance or possibility. At last, 500 identities will be sorted according to the mean value.

\begin{table}
\centering
\caption{CMC of proposed methods at Rank 1, 2, 3 on MVB.}\label{tab5}
\begin{tabular}{ C{1.5cm} C{1.5cm} C{1.5cm} C{1.5cm} C{1.5cm} C{1.5cm} C{1.5cm} }
\toprule
\multicolumn{4}{c}{Siamese Networks + } & \multirow{2}{*}{Rank1(\%)} & \multirow{2}{*}{Rank2(\%)} & \multirow{2}{*}{Rank3(\%)} \\
Merged & ATS & SE & Mask &\\
\midrule
           &            &            &            & 20.15 & 34.51 & 43.92\\
           & \checkmark &            &            & 24.24 & 39.16 & 48.00\\
           &            &            & \checkmark & 22.05 & 36.22 & 44.01\\
           & \checkmark &            & \checkmark & 26.62 & 39.26 & 47.24\\
\checkmark &            &            &            & 44.39 & 60.27 & 68.54\\
\checkmark & \checkmark &            &            & 46.39 & 58.46 & 65.49\\
\checkmark &            &            & \checkmark & 47.91 & {\bfseries61.98} & {\bfseries68.92}\\
\checkmark & \checkmark &            & \checkmark & 48.86 & 61.60 & 67.49\\
\checkmark & \checkmark & \checkmark &            & 47.72 & 59.60 & 67.30\\
\checkmark & \checkmark & \checkmark & \checkmark & {\bfseries50.19} & 61.31 & 68.73\\
\bottomrule
\end{tabular}
\end{table}

\subsection{Performance and Ablation Study}
Performance of proposed methods evaluated in form of CMC from Rank1 to Rank3 on MVB is shown in Table~\ref{tab5}. As shown, merged Siamese network shows remarkably superior results compared to basic Siamese network, ca. 20\% to 25\% boost at Rank 1, Rank 2 and Rank 3. Augmenting training set (ATS) by hard example mining can effectively improve performance, ca. 1\% to 2\% for merged Siamese network at Rank 1 and ca. 3\% to 5\% for basic Siamese network at Rank 1, Rank 2 and Rank 3. Further superior performance, i.e. 50.19\% at Rank 1 is obtained by augmenting training set and inserting SE module on masked bounding box. In real application, the most important metric is CMC Rank1, and the highest value of our baseline model is produced by combination of all model features.

\subsection{Case Study}
\begin{figure}[h]
\centering
\includegraphics[width=0.6\textwidth]{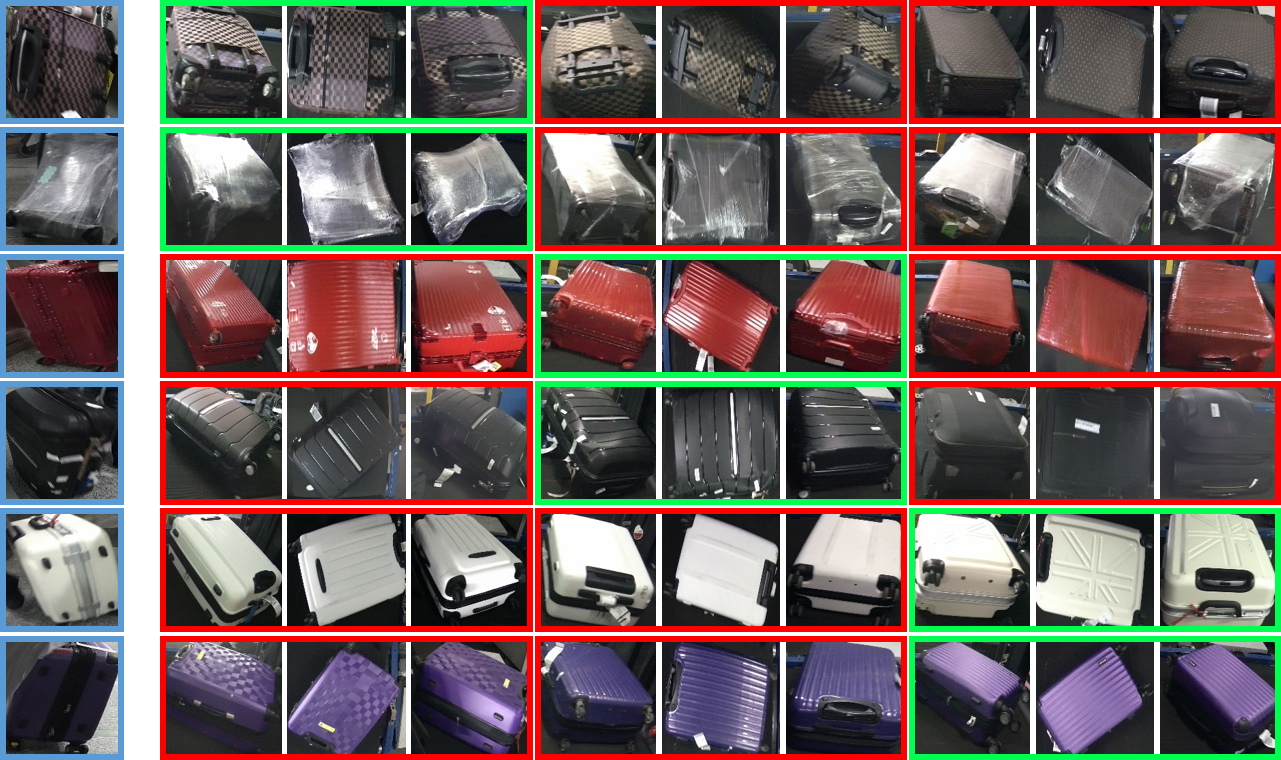}\\
(a)\\
\bigskip
\includegraphics[width=0.9\textwidth]{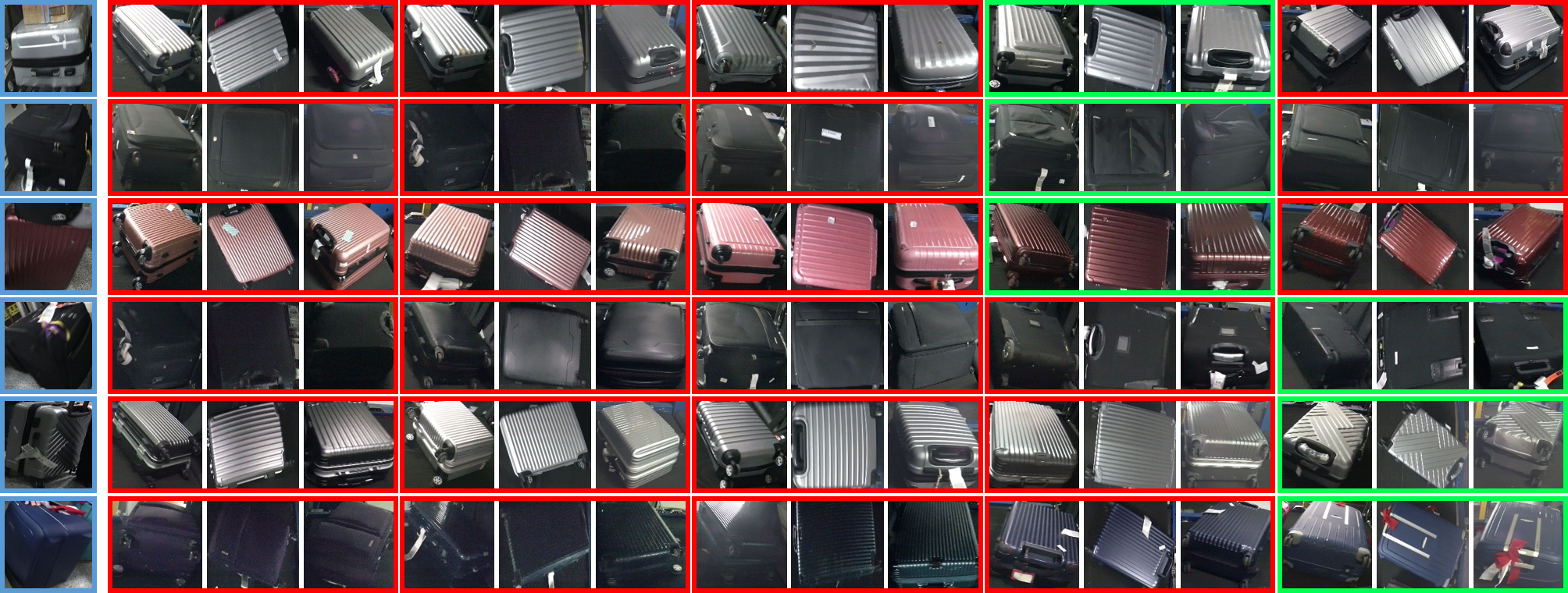}\\
(b)\\
\caption{Sample ReID results on MVB. Probe and Gallery images are not masked. Probe images are listed in the left in blue box. Gallery images are displayed in order of inferenced possibility. Gallery images with same identity as probe are bounded in green box, otherwise in red. (a) samples of baggage re-identified in top 3, (b) samples of baggage not re-identified in top 3.} \label{fig3}
\end{figure}

Sample baggage ReID results on MVB are shown in Fig.~\ref{fig3}. As shown, our proposed network can effectively retrieve baggage with similar appearance from gallery and has detail discrimination ability to some extent. Nevertheless, there are still cases where our network fails to represent more distinguishable details in retrieving baggage at top ranks. One possible reason is that our proposed network mainly extracts a global rather than local feature vector for each probe and gallery image.

\subsection{Future Work}
Baggage ReID is a research problem toward real-world application, thus the data pipeline has been set up at certain airports and will be promoted to many others. It can be expected that the scale of dataset will be continuously growing and reaching to another order of magnitude within a short period of time. Meanwhile we are organizing an open contest based on MVB for technology improvements and suggestions of dataset usage. As shown in case study, a typical mismatch is related to failing to amplify some important detail information, which is caused by the feature extraction network mainly relying on global feature. Therefore, ReID performance could be possibly improved by making better use of salient details. Last but not least, the dataset potential as 3D object ReID should be further exploited. For instance, the probe and gallery image both can be 3D image, which is baggage 3D surface reconstructed by multi camera calibration and visual SLAM; also one can apply key point detection to understand the pose of baggage, then re-identify it based on 3D alignment with some geometric shape constrains.

\section{Conclusion}
A new baggage ReID dataset named MVB is proposed in this paper. MVB consists of 4519 baggage identities and 22660 bboxes along with mask and material annotations. All data is collected in real scenario using specially-designed multi-view camera system. This paper also presented a merged Siamese network as a baseline model to work on the task of baggage ReID. Considering the large scale and the challenging factors of MVB, it will significantly contribute to further research on general 2D and 3D object ReID, especially with different domains. The performance of merged Siamese network is also evaluated as baseline model of the dataset. To access MVB dataset, please visit its corresponding contest website \href{http://volumenet.cn/}{http://volumenet.cn/}, any feedback is greatly appreciated.
%
%
%
%

\end{document}